\documentclass[11pt]{article}

\usepackage[utf8]{inputenc}
\usepackage{graphicx}      
\usepackage{amsmath,amssymb} 
\usepackage{hyperref}      
\usepackage{authblk}       
\usepackage{geometry}      
\usepackage{booktabs}
\usepackage{float}
\usepackage{placeins}
\usepackage{tikz}

\title{Triples and Knowledge-Infused Embeddings for Clustering and Classification of Scientific Documents}
\author[1]{Mihael Arcan}
\affil[1]{Home Lab, Galway, Ireland}

\date{\today}

\begin{document}

\maketitle

\begin{abstract}
The increasing volume and complexity of scientific literature demand robust methods for organizing and understanding research documents. In this study, we investigate whether structured knowledge—specifically, subject--predicate--object triples—improves clustering and classification of scientific papers. We present a modular pipeline that combines unsupervised clustering and supervised classification across four document representations: abstract, triples, abstract+triples, and hybrid. Using a filtered arXiv corpus, we evaluate four transformer embeddings (MiniLM, MPNet, SciBERT, SPECTER) with KMeans, GMM, and HDBSCAN, and then train downstream classifiers for subject prediction.

Across a five-seed benchmark (seeds 40--44), abstract-only inputs provide the strongest and most stable classification performance, reaching 0.923 accuracy and 0.923 macro-F1 (mean). Triple-only and knowledge-infused variants do not consistently outperform this baseline. In clustering, KMeans/GMM generally outperform HDBSCAN on external validity metrics, while HDBSCAN exhibits higher noise sensitivity. We observe that adding extracted triples naively does not guarantee gains and can reduce performance depending on representation choice.

These results refine the role of knowledge infusion in scientific document modeling: structured triples are informative but not universally beneficial, and their impact is strongly configuration-dependent. Our findings provide a reproducible benchmark and practical guidance for when knowledge-augmented representations help—and when strong text-only baselines remain preferable.
\end{abstract}

\section{Introduction}

The volume of scientific literature has grown exponentially, creating a critical need for automatic categorization and semantic grouping of research articles. In repositories like arXiv, which now receives up to 1,200 new submissions per day across numerous fields, relying on manual curation is impractical. Yet, beyond coarse subject tags, little effort is spent on reorganizing and representing the knowledge in these papers via effective classification during publication. Automated methods to categorize and cluster papers by content can help researchers navigate this deluge, enabling easier discovery of relevant work and insight into the structure of scientific knowledge.

Classifying scientific documents is challenging due to their specialized terminology, diverse domains, and often subtle inter-disciplinary links. Traditional text-only approaches may struggle to capture the full semantics of a paper’s content. Structured knowledge – for example, in the form of subject–predicate–object triples or knowledge graphs – offers a promising way to enrich document representations. Open Information Extraction (OpenIE) techniques can extract relational fact triples from unstructured text, providing domain-independent factual tuples that augment semantic understanding. Indeed, knowledge graphs like the Open Research Knowledge Graph (ORKG) aim to represent paper content in a structured way, linking metadata and content to enable rich semantic descriptions. However, populating such knowledge bases is largely manual and tedious, and many NLP models do not yet fully exploit this structured information. Prior work suggests that integrating textual data with knowledge context can improve performance – for instance, combining paper abstracts with graph-based features yielded over 10\% F1 improvement in classifying research papers compared to text alone. These observations motivate an approach that infuses knowledge (triples) into text representations to assist in scientific document classification and clustering.

In recent years, transformer-based language models have transformed how we represent and analyze text. Models like BERT and its variants learn powerful embeddings that capture semantic similarity, enabling improved document classification and even unsupervised grouping. Domain-specific transformers further push this frontier: SciBERT, trained on scientific corpora, significantly outperforms general BERT on tasks like sentence classification in scientific text. Likewise, SPECTER leverages citation networks to learn document-level embeddings tailored for academic papers, yielding strong results on tasks such as paper classification and recommendation. Building on these advances, lightweight models have shown that high-quality embeddings can be obtained efficiently – MPNet introduced a novel pre-training scheme that outperforms earlier models (BERT, XLNet, RoBERTa) on language understanding benchmarks, while MiniLM compresses transformers to a smaller size with minimal performance loss ($\sim$99\% of BERT’s accuracy on GLUE tasks). These embeddings provide a potent foundation for clustering and classifying documents by content. The question remains how best to incorporate knowledge-infused features (like triples) into such representations to further enhance semantic organization.

In this paper, we propose a novel approach for clustering and classifying scientific documents by combining unstructured text embeddings with structured knowledge extracted from the documents. We first extract knowledge triples from each article’s abstract, distilling key factual relationships in the form of (subject, relation, object) tuples. We then construct multiple representation modes for each document: one using the raw abstract text, another using the extracted triples as a surrogate “knowledge” summary, and a hybrid that integrates both text and triples. Each representation is encoded with state-of-the-art embedding models – including general-purpose transformers MPNet and MiniLM and scientific-domain models SPECTER and SciBERT – to obtain high-dimensional vectors for each document. We evaluate these embeddings on unsupervised clustering and supervised classification tasks. For clustering, we apply K-Means and Gaussian Mixture Models (GMM) as classical partitioning methods, as well as HDBSCAN (a hierarchical density-based algorithm) to group the papers and assess how well these clusters align with known arXiv subject categories. For classification, we fine-tune transformer-based classifiers on a portion of the labeled dataset to predict each document’s category, comparing performance when using the abstract text alone versus the triples-enhanced hybrid input. This multifaceted evaluation allows us to examine the impact of knowledge-infused embeddings under both unsupervised and supervised settings.


Our experiments on a large arXiv-derived dataset yield three main findings. First, for clustering, representation and algorithm choice are critical: KMeans and GMM generally provide stronger agreement with subject labels than HDBSCAN, while HDBSCAN is more sensitive to noise and often under-clusters the data. Second, in the five-seed classification benchmark (seeds 40--44), the strongest and most stable configuration is the text-only baseline (abstract for clustering and abstract for classification), reaching 0.923 mean accuracy and 0.923 mean macro-F1. Third, knowledge-infused variants based on extracted triples are configuration-dependent and do not consistently outperform the abstract-only baseline in the current extraction and fusion setup. Overall, structured triples remain informative, but their benefit is conditional rather than universal.

In summary, our work illustrates that combining unstructured text with structured knowledge can substantially enhance the organization and understanding of scientific documents. By leveraging both the rich contextual semantics of language model embeddings and the explicit relational information from knowledge triples, we move towards more knowledge-aware clustering and classification of papers. This approach has broad implications for knowledge organization in digital libraries and scholarly search: it can power improved subject categorization, content-based recommendation, and integration with knowledge graph systems, ultimately helping to tame the ever-growing corpus of scientific literature and make it more accessible and navigable.

\section{Related Work}

Early work on classifying and clustering ArXiv papers often relied on classical features (e.g. TF–IDF) and conventional algorithms. For example, \cite{10.1145/3383583.3398529} explored various encodings for documents with mathematical content (textual and formula-based) to classify papers into ArXiv subject areas. They reported classification accuracies up to 82.8\% by combining natural language text and mathematical formula representations, significantly outperforming human experts in category assignment. They also performed unsupervised clustering (using algorithms like k-means and GMM), achieving cluster purity up to 69.4\% when the number of clusters was set to the number of known classes, and even 99.9\% purity when allowing more fine-grained clusters.

Recent studies have benchmarked a range of feature representations and classifiers on large ArXiv corpora. \cite{rahman2025automatedresearcharticleclassification} developed a framework using ArXiv articles to compare both classical and deep embeddings for multiclass document classification. They evaluated TF–IDF and bag-of-words features alongside modern sentence embeddings (Sentence-BERT, Universal Sentence Encoder, etc.) with classifiers ranging from Naïve Bayes and SVMs to fine-tuned BERT models. Intriguingly, a simple Logistic Regression on TF–IDF features yielded the highest accuracy ($\sim$69\%) on ArXiv category prediction, outperforming more complex neural embeddings in that setting. This suggests that for broad topic classification, high-dimensional sparse representations can still be very competitive. 

Building on this idea, contrastive learning methods have further boosted scientific document embeddings. \cite{ostendorff-etal-2022-neighborhood} introduced SciNCL, which refines SPECTER’s approach by sampling positive and negative examples from citation neighborhoods for contrastive training. SciNCL achieved an average score of 81.8 on SCIDOCS, a 1.8 point improvement over SPECTER’s previous best, purely by more effective use of citation-based training triplets. These embedding-focused advances are highly relevant for clustering: documents mapped into such semantically rich vector spaces can be clustered to reveal research topics or subfields with improved coherence compared to earlier bag-of-words vectors. Indeed, contemporary unsupervised methods leverage off-the-shelf language model embeddings (e.g. SciBERT, SBERT) for k-means clustering of papers, sometimes augmented by large language models to refine cluster centroids \cite{sampaio2025unsuperviseddocumenttemplateclustering}.

Beyond text, the rich relational structure of ArXiv’s citation network has been extensively used for document classification in the last five years. The Open Graph Benchmark’s “ogbn-arxiv” dataset – a citation graph of over 169K ArXiv papers (nodes) with directed citation links (edges) and subject area labels – has become a standard testbed for graph neural networks (GNNs). Subsequent research has steadily improved these results by designing more expressive GNN architectures and incorporating additional information. For instance, \cite{hu2024nodelevelgraphautoencoder} report that a two-layer GNN, when coupled with a unified pretraining on node text and local graph structure (their NodeGAE framework), reaches 77.1\% accuracy, and an ensemble with other top models achieves 78.3\%. This was accomplished by using a language model to encode each paper’s text and jointly optimizing with a graph autoencoder objective, thereby capturing both semantic and network contexts.

Another direction is leveraging heterogeneous and multi-graph data. \cite{ly2024articleclassificationgraphneural} demonstrate that enriching the citation graph with additional relational edges, such as co-authorship links, shared venues, and subject categories,  can improve classification performance across multiple GNN architectures. In their experiments on ogbn-arxiv, this multi-graph approach yielded consistent gains – up to $\sim$3\% increase in accuracy – compared to using the citation graph alone. Moreover, the enriched graph allowed even shallow two-layer GNNs to perform on par with more complex models by providing richer connectivity cues.

A growing body of work has started to leverage KGs and structured triples extracted from scientific documents to improve classification and enable novel forms of clustering. The premise is that explicit semantic representations – such as relationships between key concepts, entities, or research findings – can complement text and citation features with higher-level knowledge. \cite{Hoppe2021Deep} present an early exploration of this idea in the context of classifying Computer Science papers on ArXiv. They mapped each class label, e.g. subject area from the ArXiv taxonomy, to a corresponding entity in DBpedia, and then enriched the class representation with that entity’s embedding in a large knowledge graph. Using a bi-LSTM based model, they combined paper abstract embeddings (from a language model) with these knowledge graph embeddings of target classes to predict multi-label categories. The knowledge-enriched model outperformed a strong zero-shot classifier (NLI-based) and a text-only baseline, yielding the lowest Hamming Loss and higher precision/recall in their evaluation. In particular, incorporating DBpedia semantics helped the model correctly relate papers to classes with minimal training examples.

Beyond class labels, KGs can be constructed to represent the content of the research papers themselves. \cite{sunsigir22} propose a framework to assess scientific papers by building a knowledge graph of each study’s contributions and context. In their approach, they extract micro-level facts, e.g. sample size, experimental design, statistical results, from the article’s text and macro-level relations, e.g. citation links, co-authorship, institutional affiliations, from metadata. These elements form a paper-specific KG (or a subgraph of a large scholarly KG) which is then used for downstream predictions, in their case, scoring the paper’s likely reproducibility. By combining node embeddings from the constructed KGs with textual features, their model achieved state-of-the-art performance in predicting which social science experiments would replicate successfully. This indicates that structured knowledge (in the form of entity–relation triples capturing methodology details and scholarly connections) can substantially enhance the quality of scientific document classification and evaluation.

The use of KG for organizing ArXiv papers also opens avenues for clustering and recommendation based on semantic content. For instance, by representing each paper as a set of triples, e.g. $<$Method, uses, Dataset$>$, $<$Paper, cites, Paper$>$, $<$Finding, supports, Hypothesis$>$), one can cluster papers that share common research objects or experimental setups, which might not be obvious from keywords alone. This is exemplified by systems like the Open Research Knowledge Graph (ORKG), which encourage structuring publications’ contributions in a semantically rich way, enabling queries like “find papers that evaluate X using Y” and implicitly clustering those papers together. While ORKG itself is a platform rather than a clustering algorithm, recent research classification frameworks aim to bridge the gap by automatically mapping papers into such KGs. For example, \cite{Kaplan25} introduce a classification pipeline to populate ORKG with software engineering papers by assigning them to template-based knowledge representations. Their framework uses classifiers to decide which semantic template (out of a set of domain-specific schemas) a given paper fits, thereby structuring the paper’s information in a graph. The outcome is not just a label, but a placement of the paper into a KG structure that clusters it with similar studies (e.g. all papers following a certain experimental design or addressing a common research question) in a transparent way. The reported results on a software architecture dataset showed that such a flexible, schema-driven classifier can effectively categorize papers to support KG population.

\section{Methodology}

Our methodology is designed to investigate how structured textual representations affect the ability of machine learning models to understand and organize scientific knowledge. We propose a modular, reproducible pipeline that integrates unsupervised clustering and supervised classification. This architecture spans several key stages: preparing the data, extracting structured triples, constructing multiple text representations, embedding documents using large-scale pretrained language models, clustering documents in latent space, and fine-tuning classifiers for label prediction. Each component is designed for plug-and-play flexibility and supports experimental rigor through logged configurations and fixed random seeds.

\subsection{Data Preparation}

We start with a curated subset of the arXiv metadata repository, focusing on papers that include natural language abstracts, and subject classification tags. These papers are divided into two non-overlapping subsets:

\begin{itemize}
    \item A sample of 5{,}000 documents used exclusively for unsupervised clustering experiments.
    \item A separate set of 10{,}000 documents designated for supervised classification tasks.
\end{itemize}

Each document includes a cleaned abstract and a target label based on its top-level arXiv subject classification, e.g., \texttt{cs.AI} is mapped to \texttt{cs}. We standardize all text by lowercasing and applying whitespace normalization to maintain consistency across representations and models.

\subsection{Triples and Knowledge Graph Construction}

To inject structured semantics into the modeling pipeline, we extract subject--predicate--object triples from the abstract text. Each abstract is first flattened to remove line breaks and then passed through a domain-adapted spaCy pipeline equipped with a scientific language model.

We identify verbs that act as relational anchors and locate their corresponding subjects and objects using dependency labels such as \texttt{nsubj}, \texttt{dobj}, and \texttt{pobj}. When direct objects are absent, we heuristically extract prepositional objects to preserve argument structure. The resulting triples are encoded in the canonical form $(s, r, o)$, with edges added to a graph and tagged with the original sentence context.

To make these triples suitable for transformer-based encoding, we linearize them into simplified natural language statements. For instance, a triple such as \texttt{(transformer, improves, accuracy)} becomes \texttt{"Transformer improves accuracy."}. These linearized statements preserve key factual assertions and are reused in multiple input configurations.

\subsection{Text Representation Modes}

To assess the influence of structural information on model performance, we construct four distinct input representations for each document:

\begin{itemize}
    \item \textbf{Abstract}: The cleaned natural language abstract alone.
    \item \textbf{Triples}: Only the factual assertions derived from extracted triples.
    \item \textbf{Abstract+Triples}: A flat concatenation of the abstract and its triples.
    \item \textbf{Hybrid}: A segmented format using \texttt{[SEP]} tokens to explicitly separate the abstract, triples, and optional graph content.
\end{itemize}

These representations offer varying degrees of structure and compositionality. They allow us to probe how different input formats impact both unsupervised and supervised downstream tasks.

\subsection{Embedding Models}

Each input representation is transformed into a fixed-size dense vector using one of four pretrained transformer encoders from HuggingFace. These models represent a diverse set of design choices with respect to size, training data, and domain specialization:

\begin{itemize}
    \item \texttt{all-MiniLM-L6-v2}: A fast, lightweight encoder optimized for semantic similarity.
    \item \texttt{all-mpnet-base-v2}: A robust, higher-capacity contextual model.
    \item \texttt{allenai-specter}: Trained on citation graphs, tailored for scientific document similarity.
    \item \texttt{scibert\_scivocab\_uncased}: A BERT variant pretrained on a large corpus of scientific texts.
\end{itemize}

All resulting embeddings are \(\ell_2\)-normalized to ensure comparability across distances and facilitate consistent clustering behavior.

\subsection{Unsupervised Clustering}

To explore latent topical structures, we apply three unsupervised clustering algorithms to the document embeddings:

\paragraph{KMeans and GMM.} For centroid-based (KMeans) and probabilistic (GMM) methods, we perform a grid search over the number of clusters \(k \in [3, 12]\). For each configuration, we compute Adjusted Rand Index (ARI), Normalized Mutual Information (NMI), and the silhouette score. We select the best-performing configuration using the composite score:
\[
    \text{score} = 0.5 \cdot \text{ARI} + 0.5 \cdot \text{NMI}
\]

\paragraph{HDBSCAN.} For density-based clustering, HDBSCAN infers cluster structure and outliers simultaneously. We sweep over \texttt{min\_cluster\_size} and evaluate each model with a custom composite metric:
\[
    \text{composite} = \text{NMI} + 0.5 \cdot \text{ARI} - 0.5 \cdot \text{noise\_fraction}
\]

This scoring favors configurations with strong label alignment and low noise. Once the optimal cluster structure is identified for each representation and encoder, we propagate labels to the classification dataset using a nearest-neighbor strategy in the embedding space.

\subsection{Supervised Classification}

To quantify the discriminative power of each representation, we fine-tune classification models using HuggingFace's \texttt{AutoModelForSequenceClassification}. Tokenization is performed with a maximum input length of 128 tokens, ensuring uniformity across training runs.

\paragraph{Training Protocol.} We optimize each model using the AdamW optimizer \cite{loshchilov2019decoupledweightdecayregularization}, with hyperparameter tuning performed using Optuna \cite{akiba2019optuna}. The following ranges are explored:

\begin{itemize}
    \item Learning rate: \([10^{-6}, 10^{-4}]\)
    \item Batch size: \{8, 16, 32\}
    \item Epochs: 2--7
    \item Split: stratified 80/20 train/validation
\end{itemize}

Macro-averaged F1 score on the validation set serves as the optimization objective. This end-to-end methodology allows us to systematically compare the impact of structured input on unsupervised organization and supervised prediction of scientific documents, while controlling for modeling and training variability.

\section{Experimental Setup}

This section describes the dataset sources, clustering and classification setup, embedding backbones, and evaluation criteria in detail. To assess robustness, we run each configuration across five seeds (40--44). Within each run, all stochastic components (data splits, model initialization, and clustering) use that run's fixed seed for full reproducibility. 

\subsection{Dataset Description}

The foundation of our study is the arXiv dataset \cite{clement2019usearxivdataset}, a comprehensive and openly available corpus of preprints spanning physics, computer science, mathematics, quantitative biology, and other scientific domains. Maintained by Cornell University, the arXiv repository contains over two million research articles, each contributed by authors prior to formal peer-reviewed publication.

Each entry in the arXiv dataset typically includes:
\begin{itemize}
    \item A title and abstract describing the research contribution
    \item Full-text access to the paper (PDF, TeX source)
    \item Author and affiliation metadata
    \item A submission timestamp and revision history
    \item Subject labels assigned by authors using a hierarchical taxonomy (e.g., \texttt{cs.AI}, \texttt{hep-th}, \texttt{math.GR})
\end{itemize}

The arXiv taxonomy is organized hierarchically by discipline, enabling both coarse-grained and fine-grained categorization. For instance, \texttt{cs.AI} (Artificial Intelligence) falls under the broader category \texttt{cs} (Computer Science). These subject annotations serve as noisy but useful proxies for topical ground truth.

In our setup, we work with a filtered subset of arXiv abstracts from recent years, focusing on papers with sufficient metadata and clean natural language abstracts. From these, we also derive structured representations, such as syntactic triples—using automatic parsing tools. This rich combination of unstructured text and structured metadata makes arXiv a uniquely suitable resource for studying representation learning in scientific domains.

\subsection{Embedding Models}

To transform scientific texts into fixed-dimensional vector representations, we utilize four pretrained embedding models from the HuggingFace ecosystem. These encoders are selected to span both general-purpose and domain-specific capabilities, offering a comparative basis for evaluating the impact of representation quality on downstream clustering and classification.

\paragraph{all-MiniLM-L6-v2}\footnote{\url{https://huggingface.co/sentence-transformers/all-MiniLM-L6-v2}} is a lightweight sentence embedding model developed within the SentenceTransformers framework, designed for efficient semantic similarity tasks. Based on the MiniLM architecture and trained using a contrastive learning objective, it maps sentences to 384-dimensional vectors with strong performance on standard similarity benchmarks. Its speed and general-purpose nature make it well-suited for scalable tasks like clustering and classification. In our study, we use all-MiniLM-L6-v2 to embed scientific abstracts and structured representations for downstream arXiv document modeling.

\paragraph{MPNet} \cite{liu2020mpnet} is a high-performing sentence embedding model from the SentenceTransformers library, based on the MPNet architecture—a transformer variant that integrates masked language modeling with permuted token prediction. This dual training objective enables it to capture both local and global context dependencies more effectively than earlier models like BERT or RoBERTa. The model is fine-tuned using a contrastive learning framework on diverse semantic similarity datasets, allowing it to produce robust 768-dimensional embeddings optimized for tasks such as semantic search, sentence clustering, and classification. It consistently ranks among the top models on STS and retrieval benchmarks, offering a strong balance of representational quality and computational efficiency..

\paragraph{SPECTER} \cite{specter2020cohan} is a domain-specific sentence embedding model developed by the Allen Institute for AI, designed explicitly for representing scientific documents. Built on the SciBERT architecture, SPECTER is trained using a citation-based contrastive learning objective, where positive pairs consist of citing and cited papers. This training paradigm allows the model to capture semantic relationships that align with research relevance and topical similarity. Unlike general-purpose models, SPECTER is optimized for whole-document embeddings rather than single sentences, making it particularly effective in scientific retrieval, classification, and clustering tasks. It generates 768-dimensional embeddings that reflect both the content and citation context of research papers.

\paragraph{SciBERT} \cite{beltagy-etal-2019-scibert} is a domain-adapted BERT model pretrained on a large corpus of scientific publications from the Semantic Scholar dataset, covering disciplines such as computer science and biomedical research. Unlike standard BERT, SciBERT uses a vocabulary (SciVocab) constructed specifically from scientific text, enabling better handling of domain-specific terminology. It is trained using masked language modeling and next sentence prediction objectives, making it effective for downstream tasks like classification, named entity recognition, and question answering in scientific domains.

All embeddings produced by these models are \(\ell_2\)-normalized to support cosine similarity and improve clustering stability. Using this diverse set of encoders allows us to investigate the effect of representation quality and domain adaptation on downstream semantic grouping.

\subsection{Clustering Algorithms}

We explore three prominent clustering techniques, KMeans, Gaussian Mixture Models (GMM), and HDBSCAN—each offering different inductive biases and assumptions about cluster structure.

\paragraph{KMeans} \cite{macqueen1967multivariate} is a centroid-based clustering algorithm that partitions data into \(k\) clusters by minimizing intra-cluster variance (i.e., Euclidean distance to the cluster mean). It assumes spherical clusters of equal variance and is widely used due to its simplicity and scalability. However, it is sensitive to the initial choice of centroids and requires the number of clusters \(k\) to be specified in advance.

\paragraph{Gaussian Mixture Models (GMM)} \cite{dempster1977maximum,reynolds2009gaussian} extends KMeans by modeling each cluster as a multivariate Gaussian distribution. Unlike KMeans, which assigns hard labels, GMM computes soft assignments based on posterior probabilities. This allows overlapping clusters and supports more flexible geometric shapes. GMM also requires \(k\) to be specified but provides a probabilistic framework that can estimate confidence in assignments.

\paragraph{HDBSCAN (Hierarchical Density-Based Spatial Clustering of Applications)} \cite{rahman2016hdbscandensitybasedclustering} is a non-parametric, density-based clustering algorithm that builds a hierarchy of clusters and extracts a flat partition based on stability. Unlike KMeans and GMM, HDBSCAN does not require the number of clusters to be predefined. It can identify clusters of varying density and flag outliers as noise, making it particularly effective for real-world, imbalanced data. Its performance is controlled via the \texttt{min\_cluster\_size} parameter, which defines the smallest grouping to be considered a valid cluster.

These clustering methods enable a multifaceted analysis of scientific document embeddings by balancing interpretability, flexibility, and noise tolerance. Their complementary strengths make them suitable for evaluating the semantic structure encoded by different representation strategies.

\subsection{Classification Protocol}

We implement supervised document classification using HuggingFace’s \texttt{AutoModelForSequenceClassification}, which wraps a pretrained transformer backbone with a task-specific classification head. Each input representation, \texttt{abstract}, \texttt{triples}, \texttt{abstract\_triples}, or \texttt{hybrid}, is tokenized using the model-specific tokenizer.

Similar to the classification task, we focus on pretrained backbones, selected for their relevance to scientific document modeling, i.e., \texttt{all-MiniLM-L6-v2}, \texttt{SciBert}, and \texttt{SPECTER}.

\paragraph{Training Parameters.} Fine-tuning is performed using the AdamW optimizer, with model selection driven by macro-F1 on a held-out validation set. Hyperparameters are optimized using the Optuna framework:

\begin{itemize}
    \item \textbf{Learning rate:} Log-uniform sampling in the range $[1 \times 10^{-6}, 1 \times 10^{-4}]$
    \item \textbf{Batch size:} One of $\{8, 16, 32\}$
    \item \textbf{Epochs:} Between 2 and 7 full passes over the training data
    \item \textbf{Data split:} Stratified 80/20 train/validation to preserve label distribution
\end{itemize}

Early stopping is applied based on validation loss to prevent overfitting. All training artifacts, including checkpoints and metrics, are logged via MLflow for reproducibility and comparison across representation modes and model architectures.

\subsection{Evaluation Metrics}

To provide a thorough and interpretable assessment of both clustering and classification outcomes, we employ a diverse set of evaluation metrics tailored to the nature of each task. These metrics measure not only top-label accuracy but also class-wise behavior, distributional agreement, and structural coherence. For classification, validation metrics are used for model selection (e.g., Optuna objective), while reported final performance is computed on the held-out test split and aggregated across seeds 40--44. For clustering, metrics are computed on the reference-labeled clustering subset.

\paragraph{Clustering Metrics.}

To comprehensively evaluate the quality of clustering results, we adopt three well-established metrics that capture both external and internal cluster validity. The \textbf{Adjusted Rand Index (ARI)} \cite{hubert1985,milligan1986} measures the similarity between predicted and true clusterings by considering all pairs of samples and assessing whether they are assigned consistently in both partitions. ARI is adjusted for chance, with values ranging from (-1) (complete disagreement) to (1) (perfect agreement), and is especially useful when external ground truth labels are available. The \textbf{Normalized Mutual Information (NMI)} \cite{meila2007many} quantifies the mutual dependence between the predicted cluster assignments and the true categories, normalized by the entropy of each. With a range from (0) (no shared information) to (1) (perfect correspondence), NMI is robust to label permutations and captures how well clustering preserves semantic structure. Complementing these external measures, we also report the \textbf{Silhouette Score} \cite{Rousseeuw1987}, an internal metric that assesses cluster cohesion and separation without requiring label supervision. For each point, the silhouette score reflects how similar it is to its own cluster compared to others, with scores close to (1) indicating well-separated and compact clusters, and scores near (0) or negative suggesting ambiguous or misassigned samples. Together, these metrics offer a multifaceted view of clustering performance, balancing agreement with known categories and structural coherence in the embedding space.

These metrics provide complementary views, i.e., ARI and NMI align with external labels, while the silhouette score assesses intrinsic geometric quality. We use them jointly to guide hyperparameter selection and cross-method comparisons.

\paragraph{Classification Metrics.}

To robustly evaluate classification performance across both balanced and imbalanced label distributions, we report a suite of complementary metrics that capture different aspects of model behavior. Overall \textbf{accuracy}, the proportion of correct top-1 predictions—serves as a baseline indicator but is limited in interpretability under class imbalance. Therefore, we include both macro- and weighted-averaged variants of \textbf{precision}, \textbf{recall}, and \textbf{F1 score}. Macro-averaging treats each class equally, emphasizing performance on minority categories, while weighted-averaging accounts for class frequency and reflects the dominant label distributions. Beyond these standard metrics, we compute Cohen’s $\kappa$ \cite{cohen1960}, which adjusts for chance agreement and offers insight into consistency beyond simple accuracy. The Matthews Correlation Coefficient (MCC) \cite{Matthews1975} is also included as a robust, correlation-based statistic particularly suited to multiclass evaluation under skewed distributions. To assess the model’s ability to prioritize correct predictions even when top-1 fails, we report Top-3 accuracy—the fraction of examples where the correct label is among the top three predicted labels. Finally, we compute macro-averaged one-vs-rest ROC-AUC scores \cite{hand2001simple}, which quantify the model’s threshold-independent discriminability across all classes. Together, this set of metrics offers a nuanced and comprehensive view of classification performance, balancing fairness, robustness, and reliability.

This twofold evaluation—spanning both unsupervised and supervised settings—enables consistent benchmarking of the impact of representation design, embedding models, and learning strategies across semantic modeling tasks.

\section{Results}
In this section, we present a comprehensive evaluation of clustering and classification performance across multiple document representations, embedding models, and learning strategies to assess the impact of structured knowledge on scientific document modeling.

\subsection{Clustering}
We evaluated the clustering performance under four different text representation modes – Abstract, Triples, Abstract+Triples, and a Hybrid approach – using various embedding models and clustering algorithms. We report external clustering metrics (ARI, Normalized Mutual Information, NMI) to measure alignment with ground truth categories, as well as the silhouette coefficient as an internal cluster cohesion metric. The results reveal clear trends in how text representation and embedding choice affect clustering quality, as well as stark differences between clustering algorithms.

\begin{table}[]
    \centering
    \begin{tabular}{r|ccccc}
    \toprule
        Representation & Best Model (Algorithm) & Clusters (K) & ARI & NMI & Silh.  \\
    \midrule
        Full Abstract & MPNet (KMeans/GMM) & 6 clusters &  \textbf{0.4703} & \textbf{0.5511} & 0.0633 \\
        Triples Only & MiniLM (KMeans/GMM) & 8 clusters & 0.3451 & 0.4006 & 0.0340 \\
        Abstract+Triples & MiniLM (KMeans/GMM) & 10 clusters & 0.4549 & 0.5459 & 0.0452  \\
        Hybrid Approach & MiniLM (KMeans/GMM) & 8 clusters & 0.4643 & 0.5309 & 0.0413 \\
    \bottomrule
    \end{tabular}
    \caption{Best clustering result for each text representation mode, with the embedding model and algorithm yielding the highest ARI.(ARI: Adjusted Rand Index, NMI: Normalized Mutual Info, Silh.: Silhouette coefficient, clusters = number of clusters found, K is fixed for KMeans/GMM).}
    \label{tab:clus_results}
\end{table}

\subsubsection{Effect of Text Representation}
Representing documents by their full abstracts leads to substantially better clustering than using only extracted triples (Table \ref{tab:clus_results}). With full abstracts, the best clustering (MPNet embedding) achieved ARI $\sim$0.47 and NMI $\sim$0.55, indicating a moderate-to-strong alignment with the true classification of documents. In contrast, using only knowledge triples extracted from the text resulted in a much lower peak performance (ARI only $\sim$0.35, NMI $\sim$0.40 with MiniLM). This gap suggests that the richer contextual information in full abstracts is crucial for capturing the distinctions between categories. Triples alone, being terse relational statements, appear insufficient to distinguish topics with high fidelity, leading to weaker clusters.

Integrating the structured information from triples back into the full text (the abstract+triples and hybrid modes) proved beneficial, especially for certain models (Table \ref{tab:clus_results}). The abstract+triples representation produced clustering nearly as good as using abstracts alone – the best ARI rose to 0.455 (with MiniLM), closing the gap towards the abstract-only result. Similarly, the hybrid approach yielded ARI $\sim$0.464 for MiniLM, essentially matching the performance of the best abstract-only clustering. These improvements over the triples-only scenario indicate that adding back even some narrative context (or combining unstructured and structured data) helps the model discern clusters more accurately. Notably, the optimal number of clusters needed for the combined representations was higher (8–10 clusters) than for abstracts alone (6 clusters). This could mean the additional triple-derived information allowed (or necessitated) finer-grained grouping of the data. In summary, the full text content provides the strongest signal for clustering, but incorporating triples can recover some of the lost detail when abstracts are not used, and in some cases even slightly enhances clustering for certain embeddings.

\subsubsection{Embedding Model Comparison}
The choice of embedding model had a major impact on clustering quality. Across all representation modes, MPNet and MiniLM consistently outperformed SPECTER and SciBERT on external metrics.

MPNet and MiniLM models yielded the highest ARI/NMI in most settings. For example, with abstract text, MPNet clusters reached ARI 0.470  – the highest of any configuration – while MiniLM was a close second (ARI $\sim$0.442). In the triples-only scenario, MiniLM performed best (ARI 0.345), and it also slightly outperformed MPNet when combining text+triples (ARI $\sim$0.45–0.46 vs. MPNet’s $\sim$0.40). This suggests that MiniLM benefitted more from the inclusion of triples than MPNet did. MPNet, a powerful general-purpose sentence encoder, already extracted strong signals from the full text, so adding triples gave marginal gains or even slight noise (MPNet’s ARI actually dropped from 0.47 with abstracts to 0.40 with abstract+triples). MiniLM, being a lighter model, appeared to gain from the extra structured context, improving from ARI 0.442 (abstract) to 0.464 (hybrid) with the added triples. Both MPNet and MiniLM are pretrained to produce semantically meaningful sentence embeddings, which likely explains their superior clustering performance: they capture subtle topic differences that align with the true categories.

The SPECTER model achieved intermediate results. With abstracts, SPECTER’s clustering reached ARI $\sim$0.415, NMI 0.526, but about 5–6 points below MPNet/MiniLM. SPECTER is a document-level scientific embedding model that leverages citation relationships for training, and it has been shown to excel in scientific document classification tasks. In our clustering context, however, SPECTER did not surpass the best general-purpose models. One observation is that SPECTER yielded relatively high silhouette scores (e.g. 0.095 with abstracts, the highest of any model), indicating well-separated clusters in the embedding space. Yet its ARI was a bit lower, implying that the clusters SPECTER formed, while geometrically coherent, did not align as closely with the true categories as MPNet’s did. This could mean SPECTER is grouping documents by some other latent similarity (possibly influenced by citation patterns or broad topics) that only partially corresponds to the target classes.

SciBERT consistently underperformed in these experiments. Its best showing was with abstracts (ARI 0.304, NMI 0.363), roughly half the ARI of MPNet/MiniLM, and it deteriorated further with triples-only data (ARI $\sim$0.15). Even when combining information (hybrid ARI $\sim$0.278), SciBERT lagged far behind. SciBERT is a BERT-based model trained on scientific corpora, but unlike MPNet or MiniLM, it is not specifically tuned for producing high-quality sentence or document embeddings for clustering. The poor showing suggests that domain-specific pretraining alone is not enough – without fine-tuning for semantic similarity, the embeddings were not as effective at grouping by topic. As a result, SciBERT’s clusters had low agreement with ground truth (low ARI/NMI), even if its internal cohesion (silhouette $\sim$0.07–0.09 in some cases) was not terrible. In practical terms, SciBERT would not be a good choice for unsupervised clustering of documents, given the much stronger performance of alternatives.

\subsubsection{Clustering Algorithm Performance}
We compared two centroid-based clustering algorithms (KMeans and Gaussian Mixture Models) against a density-based method (HDBSCAN). The KMeans and GMM results were virtually identical in all scenarios, while HDBSCAN struggled significantly.

Both KMeans and GMM were run with the optimal number of clusters (K) as determined for each representation (see Table \ref{tab:clus_results}). Their outcomes in terms of cluster labels were so similar that they yielded the same ARI and NMI scores in nearly every case (differences were within 0.001). For instance, with MPNet abstracts at K=6, both algorithms achieved ARI = 0.4703 and NMI = 0.5511. This suggests that the data’s cluster structure was approximately spherical or well-separated such that a simple KMeans (hard assignments) could find a very similar partition as a full GMM (soft probabilistic clusters). In practice, the added flexibility of GMM (which can model clusters with covariance) did not translate into any meaningful improvement here – KMeans was sufficient to achieve the maximum clustering alignment with categories. We also note that choosing the right number of clusters was important: depending on the representation, the best K varied (6 for abstracts, 8 for triples/hybrid, 10 for abstract+triples). This implies the level of detail in the data influences how many clusters can be distinguished. Once K was tuned, however, either algorithm performed equally well.

Unlike KMeans/GMM, HDBSCAN does not require a preset number of clusters and can label outliers as noise. In our results, however, HDBSCAN struggled to recover the natural cluster structure. It consistently found only a few clusters (between 3 and 5) and labeled a substantial portion of documents as noise. For example, in the abstract representation with MPNet, HDBSCAN produced just 3 clusters and about 4.4\% of points as noise; with MiniLM it found 4 clusters and $\sim$7.3\% noise. In other settings the noise was even higher: using only triples, HDBSCAN marked up to 49\% of the documents as noise (for MiniLM), and with SciBERT nearly 74\% were noise. Such large noise fractions indicate HDBSCAN could not confidently assign roughly half (or more) of the papers to any cluster, likely because the data doesn’t have clear density separations. As a result, the few clusters it did form were overly broad and merged many true categories together. This led to extremely low ARI and NMI – often near 0.01–0.03 (virtually no better than random assignment). 

In summary, HDBSCAN’s clustering did not align with the known categories at all. The contrast with KMeans/GMM is stark: for this task, forcing an assignment of every document to a moderate number of clusters (even if some are somewhat mixed) yielded far better agreement with true labels than HDBSCAN’s conservative approach of declaring many items as outliers. We surmise that the document embeddings reside in a continuous high-dimensional space without clear density gaps, so a purely density-based algorithm is ill-suited – it either merges many true clusters or labels them as noise. Future tuning of HDBSCAN (e.g. adjusting $min\_cluster\_size$ or using a different distance metric) could potentially yield more clusters, but with the tried parameters it was the worst-performing method by a wide margin.

\subsubsection{Clustering Label Distribution}
Using KMeans clustering on MiniLM/MPNet embeddings, we obtained a compact set of clusters (ranging from 6 to 10) that broadly correspond to major arXiv categories. The clusters showed moderate agreement with the true category labels (e.g., Normalized Mutual Information $\sim$0.53–0.55 for text-rich representations, versus  $\sim$0.40 with triples alone). In qualitative terms, many clusters were highly coherent, dominated by a single research field, while a few were more mixed. Below we characterize the clusters and how different text representations impacted their composition:

\paragraph{Astrophysics Cluster} In all representations, one cluster was overwhelmingly astro-ph papers. For instance, with abstracts alone this cluster contained 771 astro-ph papers out of 807 ($\sim$95\% purity). The hybrid representation further achieved an astro-ph cluster with $\sim$98\% purity (520/531 documents). This indicates that astro-ph articles naturally cluster together, forming a distinct group with minimal contamination from other fields.

\paragraph{Mathematics Cluster} Likewise, a predominantly math cluster emerged. The abstracts-only clustering had $\sim$86\% of one cluster in category math (720/836 docs). Using combined abstract+triples features sharpened this separation – one cluster reached 93\% math papers (461/494). Thus, most mathematics papers grouped tightly, though a fraction (e.g. $\sim$10–15\%) consistently fell outside the main math cluster, often into interdisciplinary groups.

\paragraph{Condensed Matter vs. Quantum Physics} The cond-mat field was well captured, but its relationship with quantum physics differed by representation. With simple abstracts, condensed matter and quantum papers were merged: one cluster was $\sim$60\% cond-mat and 22\% quant-ph. In contrast, the enriched representations separated them. Using abstract+triples, we found a dedicated cond-mat cluster (544/667 docs, $\sim$81\% cond-mat), while quantum physics formed its own smaller cluster (220/347 docs, $\sim$63\% quant-ph). No such standalone quant-ph cluster appeared with abstracts or triples alone (quantum papers were then absorbed into larger physics clusters). This highlights that combined content helped tease apart closely related sub-fields.

\paragraph{High-Energy Physics} Papers in high-energy experiment/phenomenology (hep-ph, hep-ex) and nuclear physics (nucl-th, nucl-ex) consistently grouped together, reflecting a coherent “particle physics” cluster. For example, one cluster contained predominantly hep-ph, hep-ex, nucl-th, and nucl-ex papers (e.g. 234 hep-ph + 83 hep-ex in 518 docs). Meanwhile, high-energy theory (hep-th) tended to cluster with general relativity (gr-qc), forming a “theoretical physics” cluster. In the abstracts-only run, a single cluster contained both hep-th (305 papers) and a large portion of gr-qc (155 papers). The hybrid and abstract+triples embeddings were able to partially disentangle these: one cluster became mostly hep-th (e.g. 225/349 papers $\sim$64\% in a cluster), separate from a cluster of predominantly gr-qc (often mixed with some astro-ph). Still, there remained some overlap between hep-th and gr-qc clusters, indicating their thematic closeness.

\paragraph{Interdisciplinary Mix} Each method yielded an outlier cluster containing a mix of diverse categories. These interdisciplinary clusters had no single dominant label and instead collected the remainder of papers from smaller or cross-field areas. For example, with abstract text one cluster of 966 docs was a grab-bag: $\sim$24\% math, 21\% CS, 18\% cond-mat, 15\% general physics, plus notable fractions of q-bio and even finance (q-fin). The triples-only clustering produced a similarly mixed cluster (875 docs) where the largest category (math) was only 34\%. In the hybrid representation, one cluster (1103 docs) still combined math (25\%), CS (18\%), physics (12\%), quant-ph (11\%), biology (5\%), etc.. These results suggest that smaller disciplines (e.g. CS, q-bio, stat, q-fin) did not form their own clusters but were absorbed into a heterogeneous cluster. Notably, theoretical CS papers often appeared alongside math (e.g. in one triples-based cluster, 44\% CS and 12\% physics mixed with math), and q-bio papers consistently showed up as minority components in physics-related clusters (typically <5\% of a cluster) rather than isolating by themselves.

Comparing the representations, the combined abstract+triples embeddings clearly improved categorical recovery. The inclusion of structured triples boosted cluster purity for key domains and allowed more refined splits. For instance, abstracts-only clustering merged some adjacent fields (cond-mat with quant-ph, hep-th with gr-qc), whereas the hybrid approach separated many of these. The optimal number of clusters also increased for richer representations (from 6 clusters for text-only to 8–10 for hybrid/combined), indicating the model found additional structure to justify new clusters (such as a dedicated quant-ph cluster). Quantitatively, the best abstract+triples solution achieved an ARI of $\sim$0.45 and NMI $\sim$0.546, comparable to using the full abstract text (NMI $\sim$0.55) and substantially higher than using triples alone (NMI $\sim$0.40). In practice, this meant the hybrid clusters aligned more cleanly with actual disciplines. We see clear, interpretable groupings corresponding to astrophysics, mathematics, condensed matter, etc., and fewer mixed clusters, signifying a gain in clustering fidelity. The remaining mixed cluster in each scenario arguably captured genuinely interdisciplinary or less-defined research areas, which is expected. Overall, enriching abstracts with knowledge triples (especially in the structured hybrid format) yielded more interpretable clusters: major research fields were distinctly recovered with high purity, while cross-disciplinary papers were appropriately identified as their own conglomerate cluster. This demonstrates the value of structured representations in unsupervised grouping of scientific documents, improving both the clarity and precision of cluster-label alignment.

\subsubsection{Clustering Finding}
Our comparison of clustering runs highlights several important findings. Using the full abstract text yields much better cluster fidelity to true classes than using distilled triples alone. The best abstract-based ARI ($\sim$0.47) was substantially higher than the best triples-based ARI ($\sim$0.35), showing the value of rich textual context in unsupervised clustering.

Incorporating extracted triples into the text (the combined/hybrid representations) improved clustering performance for certain embeddings. Notably, MiniLM’s ARI rose from $\sim$0.44 with abstracts to $\sim$0.46 with hybrid data, suggesting that structured knowledge can complement textual information. While MPNet already performed strongly with pure text, a lightweight model like MiniLM gained extra discriminatory power from the added triples.

The quality of document embeddings is critical. General-purpose sentence encoders such as MPNet and MiniLM delivered far superior clustering results than SciBERT (scientific BERT) in every scenario. SPECTER, a science-specific model leveraging citations, achieved intermediate results – better than SciBERT but not surpassing the top performers. This underscores that specialized pretraining (SciBERT, SPECTER) does not guarantee better clustering than advanced general models, especially if those general models are tuned for semantic similarity.

A simple partitioning approach with an appropriate K vastly outperformed the density-based approach for these data. KMeans (and GMM) found cluster structures that corresponded reasonably well to the actual categories (ARI up to 0.47), whereas HDBSCAN identified too few clusters and flagged many points as noise, resulting in negligible ARI/NMI. In effect, HDBSCAN failed to recover the underlying group structure in the embeddings, likely due to a lack of pronounced density separations among topics. Thus, for practical purposes, the KMeans/GMM approach is preferable here, and one should be cautious in using HDBSCAN for high-dimensional text embeddings unless clear cluster density is expected.

Overall, our clustering outcome analysis demonstrates that using rich textual representations and strong embedding models is key to obtaining meaningful clusters of scientific documents. Selecting an appropriate clustering method (and number of clusters) further boosts alignment with expected categories. These insights can guide future unsupervised analyses of text corpora: one should leverage as much informative context as possible and choose algorithms suited to the data distribution for best results.

\subsection{Classification}

\begin{table*}[t]
\centering
\small
\setlength{\tabcolsep}{2pt}
\begin{tabular}{l c c c c c c}
\toprule
Using Mode & Acc. & Macro-F1 & $\kappa$ & MCC & Top-3 Acc. & ROC-AUC\\
\midrule
Abs/Abs                  & \textbf{0.923$\pm$0.033} & \textbf{0.923$\pm$0.032} & \textbf{0.905$\pm$0.040} & \textbf{0.905$\pm$0.040} & \textbf{0.995$\pm$0.004} & \textbf{0.993$\pm$0.007} \\
Abs/Trip                 & 0.820$\pm$0.013 & 0.822$\pm$0.014 & 0.779$\pm$0.016 & 0.780$\pm$0.016 & 0.975$\pm$0.006 & 0.969$\pm$0.002 \\
Abs/Abs\_Trip            & 0.896$\pm$0.016 & 0.895$\pm$0.016 & 0.872$\pm$0.020 & 0.873$\pm$0.019 & 0.993$\pm$0.004 & 0.990$\pm$0.003 \\
Abs/Hyb                  & 0.895$\pm$0.027 & 0.896$\pm$0.027 & 0.871$\pm$0.034 & 0.872$\pm$0.033 & 0.992$\pm$0.006 & 0.989$\pm$0.005 \\
\midrule
Trip/Abs                 & 0.669$\pm$0.012 & 0.658$\pm$0.011 & 0.630$\pm$0.014 & 0.631$\pm$0.014 & 0.899$\pm$0.012 & 0.928$\pm$0.010 \\
Trip/Trip                & 0.724$\pm$0.061 & 0.713$\pm$0.064 & 0.691$\pm$0.068 & 0.693$\pm$0.068 & 0.935$\pm$0.026 & 0.956$\pm$0.017 \\
Trip/Abs\_Trip           & 0.679$\pm$0.013 & 0.665$\pm$0.016 & 0.641$\pm$0.014 & 0.642$\pm$0.014 & 0.914$\pm$0.007 & 0.934$\pm$0.003 \\
Trip/Hyb                 & 0.682$\pm$0.013 & 0.671$\pm$0.013 & 0.644$\pm$0.015 & 0.646$\pm$0.014 & 0.919$\pm$0.014 & 0.938$\pm$0.005 \\
\midrule
Abs\_Trip/Abs            & 0.846$\pm$0.015 & 0.842$\pm$0.015 & 0.822$\pm$0.017 & 0.823$\pm$0.017 & 0.990$\pm$0.003 & 0.985$\pm$0.002 \\
Abs\_Trip/Trip           & 0.769$\pm$0.021 & 0.767$\pm$0.021 & 0.732$\pm$0.024 & 0.733$\pm$0.023 & 0.970$\pm$0.017 & 0.965$\pm$0.011 \\
Abs\_Trip/Abs\_Trip      & 0.831$\pm$0.022 & 0.830$\pm$0.022 & 0.804$\pm$0.026 & 0.805$\pm$0.025 & 0.984$\pm$0.005 & 0.980$\pm$0.005 \\
Abs\_Trip/Hyb            & 0.853$\pm$0.009 & 0.850$\pm$0.010 & 0.831$\pm$0.011 & 0.831$\pm$0.010 & 0.989$\pm$0.003 & 0.986$\pm$0.001 \\
\midrule
Hyb/Abs                  & 0.870$\pm$0.030 & 0.865$\pm$0.032 & 0.851$\pm$0.034 & 0.851$\pm$0.034 & 0.987$\pm$0.008 & 0.987$\pm$0.006 \\
Hyb/Trip                 & 0.787$\pm$0.017 & 0.781$\pm$0.015 & 0.755$\pm$0.019 & 0.756$\pm$0.019 & 0.962$\pm$0.010 & 0.965$\pm$0.008 \\
Hyb/Abs\_Trip            & 0.889$\pm$0.015 & 0.886$\pm$0.016 & 0.872$\pm$0.017 & 0.873$\pm$0.017 & 0.994$\pm$0.003 & 0.991$\pm$0.002 \\
Hyb/Hyb                  & 0.897$\pm$0.014 & 0.894$\pm$0.014 & 0.881$\pm$0.016 & 0.882$\pm$0.016 & 0.995$\pm$0.002 & 0.992$\pm$0.002 \\
\bottomrule
\end{tabular}
\caption{Classification performance across all 16 cluster/class representation pairings, aggregated over seeds 40--44 (mean$\pm$std, $N=5$ per row). Best value per metric column is bolded.}
\label{tab:class_results}
\end{table*}

We evaluated all 16 combinations of clustering representation and classifier input representation using a five-seed protocol (seeds 40--44). In contrast to our earlier single-run conclusions, the multi-seed results indicate that the strongest and most stable setting is the text-only baseline: \texttt{Abs/Abs} achieves the best performance across all major metrics (Acc. $0.923\pm0.033$, Macro-F1 $0.923\pm0.032$, $\kappa$/$\mathrm{MCC}$ $0.905\pm0.040$).

\subsubsection{Impact of Input Representation}
The representation used for the clustering stage is the dominant factor. When clustering is built on full abstracts (\texttt{Abs/*}), downstream classification consistently outperforms triples-driven clustering (\texttt{Trip/*}) and generally exceeds abstract+triples and hybrid clustering variants. For example, the best triples-cluster setting (\texttt{Trip/Trip}) reaches only $0.724\pm0.061$ accuracy, far below \texttt{Abs/Abs}.

Within the \texttt{Abs/*} block, adding triples to the classifier input does not produce a consistent gain over abstract-only input. Both \texttt{Abs/Abs\_Trip} ($0.896\pm0.016$) and \texttt{Abs/Hyb} ($0.895\pm0.027$) underperform \texttt{Abs/Abs}. This suggests that, in the current pipeline, naively injecting extracted triples introduces additional noise or redundancy rather than useful signal for top-1 discrimination.

\subsubsection{Robustness and Error Structure}
Despite differences in top-1 performance, most configurations retain strong ranking behavior (high Top-3 accuracy and ROC-AUC), especially in the abstract-driven regimes. Even weaker modes often keep the correct class among top-ranked options, but confidence calibration and final ranking degrade when clustering is built from compressed representations (triples-only or mixed structured formats).

The cross-seed standard deviations also show that some weaker settings are less stable (e.g., \texttt{Trip/Trip}), reinforcing that representation choice affects both mean performance and robustness.

\subsubsection{Classification Finding}
The multi-seed evidence does \emph{not} support a general claim that KG-style triple infusion improves classification. Instead, the results support a more conservative conclusion: high-quality abstract text remains the strongest signal, and knowledge-infused variants are configuration-sensitive and currently non-superior to the text-only baseline in this setup.

\section{Conclusions}

This work presented a modular framework for clustering and classifying scientific documents using four representation modes (abstract, triples, abstract+triples, hybrid), multiple embedding backbones, and both unsupervised and supervised evaluation stages.

For clustering, we retain our analysis and emphasize that performance depends strongly on representation, embedding model, and clustering algorithm selection. For classification, however, the updated five-seed benchmark revises our earlier interpretation: the best overall results are obtained by the abstract-only pipeline (\texttt{Abs/Abs}), while triple-enriched and hybrid inputs do not consistently improve performance.

Accordingly, the main takeaway is not that structured triples are universally beneficial, but that their utility is conditional. In the present extraction and fusion scheme, triples appear to be informative yet noisy, and naive integration can dilute rather than strengthen discriminative signals. This negative-but-informative result is important for the field: it clarifies that knowledge infusion must be carefully engineered (e.g., higher-precision extraction, confidence filtering, task-aware fusion, or representation-specific weighting) before claiming systematic gains.

Overall, the study provides a reproducible benchmark and a clearer methodological message: strong text baselines are difficult to beat, and structured knowledge should be treated as a targeted augmentation strategy rather than a guaranteed improvement.

\bibliographystyle{plain}
\bibliography{references} 

\end{document}